\documentclass[journal]{IEEEtran}
\usepackage{amsfonts}
\usepackage{pseudocode}
\usepackage{algorithmicx}
\usepackage{algorithm,algpseudocode}
\usepackage{xcolor}

\usepackage{cite}

\usepackage{pgfplots}
\usepackage{tikz}
\pgfplotsset{compat=newest}
\usetikzlibrary{plotmarks}
\usetikzlibrary{arrows.meta}
\usepgfplotslibrary{patchplots}
\usepackage{grffile}
\usepackage{amsmath}

\usepackage{amsmath,amssymb,amsthm}
\usepackage{graphicx}
\usepackage{subfigure}
\usepackage{color}
\usepackage{graphics}
\usepackage{verbatim,bm}
\usepackage{setspace}
\usepackage{enumitem}
\usepackage{bbm}

\usepackage{booktabs}
\usepackage{caption}
\title{
	A Distributed Model-Free Algorithm for Multi-hop Ride-sharing using Deep Reinforcement Learning  
}

\author{Ashutosh Singh, Abubakr Alabbasi, and Vaneet Aggarwal% 
	\thanks{The authors are with  Purdue University, West Lafayette IN 47907, email: \{singh596,aalabbas,vaneet\}@purdue.edu.}
		\thanks{Part of this work is accepted for publication in the Proceeding of Neurips Workshop, Dec 2019 \cite{MHRS_neurips} (to appear).}
}

\if 0
\author{%
  Happy Horizontal People Transporter\thanks{Use footnote for providing further information
  about author (webpage, alternative address)---\emph{not} for acknowledging funding agencies.} \\
  Genuine People Personality Division\\
  Sirius Cybernetics Corporation\\
  \texttt{marvin@siriuscybernetics.com} \\
}
\fi

\begin{document}
\maketitle

\begin{abstract}
  The growth of autonomous vehicles, 
  ridesharing systems, and self driving technology will bring a shift in the way ride hailing platforms plan out their services. However, these advances in technology coupled with road congestion, environmental concerns, fuel usage, vehicles emissions, and the high cost of the vehicle usage have brought more attention to better utilize the use of vehicles and their capacities.
   In this paper, we propose a novel multi-hop ride-sharing (MHRS) algorithm that uses deep reinforcement learning to learn optimal vehicle dispatch and matching decisions by interacting with the external environment. By allowing customers to transfer between vehicles, i.e.,
ride with one vehicle for sometime and then transfer to another one, MHRS helps in attaining 30\% lower cost and 20\% more efficient utilization of fleets, as compared to the ride-sharing algorithms. This flexibility of multi-hop feature gives a  seamless experience to customers and ride-sharing companies, and thus improves ride-sharing services.
\end{abstract}

\begin{IEEEkeywords}
Multi-hop, Ride-sharing,  Passenger transfer, Matching, Deep Q-network, Vehicle dispatch, Distributed algorithms.
\end{IEEEkeywords}

% \documentclass{article}
% % if you need to pass options to natbib, use, e.g.:
% %     \PassOptionsToPackage{numbers, compress}{natbib}
% % before loading neurips_2019

% % ready for submission
% \usepackage{neurips_2019_ml4ad}

% % to compile a preprint version, e.g., for submission to arXiv, add add the
% % [preprint] option:
% % \usepackage[preprint]{neurips_2019_ml4ad}

% % to compile a camera-ready version, add the [final] option, e.g.:
% % \usepackage[final]{neurips_2019_ml4ad}

% % to avoid loading the natbib package, add option nonatbib:
% % \usepackage[nonatbib]{neurips_2019_ml4ad}

% \usepackage[utf8]{inputenc} % allow utf-8 input
% \usepackage[T1]{fontenc}    % use 8-bit T1 fonts
% \usepackage{hyperref}       % hyperlinks
% \usepackage{url}            % simple URL typesetting
% \usepackage{booktabs}       % professional-quality tables
% \usepackage{amsfonts}       % blackboard math symbols
% \usepackage{nicefrac}       % compact symbols for 1/2, etc.
% \usepackage{microtype}      % microtypography
% \usepackage{graphicx}
% \graphicspath{ {./images/} }
% \begin{document}

% \begin{abstract}
%   The abstract paragraph should be indented \nicefrac{1}{2}~inch (3~picas) on
%   both the left- and right-hand margins. Use 10~point type, with a vertical
%   spacing (leading) of 11~points.  The word \textbf{Abstract} must be centered,
%   bold, and in point size 12. Two line spaces precede the abstract. The abstract
%   must be limited to one paragraph.
% \end{abstract}

\section{Introduction}

\subsection{Motivation}

As the adoption of autonomous vehicles becomes more widespread, the need for ride-sharing and carpooling transportation will become even more prominent. Autonomous driving could lead to fewer accidents, fewer traffic deaths, greater energy efficiency, and lower insurance premiums \cite{litman2017autonomous,anderson2014autonomous,blyth2016expanding}. It is estimated that savings with autonomous vehicles for the United States alone will be around \$1.3 trillion \cite{kearney2016automakers}. Carmakers race to develop self-driving cars to use in fleets of robo-taxis to tap into the potentially lucrative new market. While 
ride-sharing (joint travel of two or more passengers in a single vehicle) has long been a common way to share the costs and reduce the congestion, it remains a nook transportation option with
limited route choice  and
sparse schedules (only few rides per route).
This work explores the benefits of allowing customers to transfer between vehicles to further improve the ride-sharing services. 

%
%There is a need of such technologies that reduce the congestion on roads, but at the same time provide a seamless ride experience to consumers. 
Multi-hop ride-sharing (MHRS) plays a crucial role in revolutionizing the transport experience in ride-sharing and logistics transportation \cite{drews2013multi,teubner2015economics,chen2019pptaxi}. 
With the advancement of self-driving vehicle technology and proliferation of ride-sharing apps (e.g., Uber and Lyft), optimal distributed dispatching of vehicles is essential to minimize the supply-demand mismatch, reduce the waiting time of customers, reduce the travel cost, and utilize the fleet effectively. Also, multi-hop ride-sharing is a more competitive and reliable mode of transportation in terms of time and cost when compared with other modes of transport such as trains and buses \cite{coltin2014ridesharing,Teubner2015}. However, hopping and dispatch decisions over a large city for carpooling requires instantaneous decisions for thousands of drivers to serve the ride requests over an uncertain future-demand.
Moreover, the dispatch decision for each vehicle could be time consuming, exacerbating the uncertainty of the optimization over a dynamically changing demand. We note that vehicle can only be dispatched if it is not  fully occupied, and the travel time of a passenger, if any, within the vehicle should be minimized. Thus, the dispatch decision depends on the future potential trip requests which are uncertain in nature. Yet, realistic models are necessary to assess the trade-offs between possibly conflicting interests, minimizing the un-served ride requests, waiting time per request, total trip times of passengers, and the total trip distance of every vehicle.

One of the necessary components for a successful dynamic MHRS system is an efficient optimization techniques that matches drivers and riders in
real-time. However, the majority of work in this domain (e.g., \cite{zhu2016public,agatz2012optimization,greenwood2017show}) requires pre-specified model for demand prediction,  user's utilities for waiting times and travel times, and cost functions governing the cars' total trip distance. These impact the decisions of route planning for the vehicles in real-time. Thus, a model based approach is unable to consider all the intricacies involved in the optimization problem. Further, a distributed algorithm is preferable where each vehicle can take its own decision without coordinating with the others. This paper aims to consider a model-free distributed approach for ride-sharing with carpooling with possible passenger transfer from one vehicle to another one. Our approach can adapt to the changing distributions of customer arrivals, the changing distributions of the active vehicles, the vehicles' locations, and user's valuations for the total travel time. The tools from deep reinforcement learning \cite{mnih2015human} are used for this dynamic modeling, where the vehicle dispatching is dynamically learned using the deep neural network, based on which the action is chosen with the Q-learning strategy. 

%rideshare services which transfer passengers
%between multiple drivers

\subsection{Related Work}
Multi-hop ride-sharing poses myriad challenges that need to be addressed for efficient ride platforms services \cite{drews2013multi,teubner2015economics,sonet2019shary}. While fleet management and ride-sharing problems have been widely studied, most of the approaches in the literature are model-based, see \cite{zheng2017online,zhang2016control,ma2015real,gopalakrishnan2016costs,bei2018algorithms,chen2019pptaxi,masoud2017decomposition} and the references therein. In such models, 
pickup request locations, travel time, demand prediction, route decisions, and destinations are assumed to follow certain distributions and then propose dispatching policies that would improve the performance \cite{coltin2014multi,coltin2013towards}.
Using realistic data from ride-sharing services, the authors in \cite{jauhri2017space} have modeled the customer requests and their variations over service area using a graph. The  temporal/spatial variability of ride requests and the potentials for ride-pooling are captured. In \cite{de2015comewithme}, a matching algorithm based on the utility of the desired user is proposed.
However, the proposed approach is a model-based that lacks adaptability and thus becomes impractical when the number of vehicles is very large, which is the scenario in our setting. 
%In \cite{bistaffa2017cooperative}, a cooperative game theoretic approach is proposed to study the ride-sharing problem. The problem is formulated as a graph-constrained coalition formation (GCCF) and then used CFSS algorithm to solve it.  The ridesharing problem is also cast under the pickup and delivery problem (PDP), e.g., \cite{dumitrescu2010traveling,furuhata2013ridesharing}. Authors in \cite{lu2015optimization} have optimized the number of miles driven by
%drivers by pooling riders; ride requests are generated uniformly over square blocks. Based on analyzing
%real-world data from a ride-sharing service, the authors in \cite{jauhri2017space} have modeled the ride requests and their variations over location and time using a graph. The spatial and temporal variability of ride requests and the potentials for ride pooling are captured.

Recently, different approaches for dispatching vehicles and allocating transportation resources in modern cities are investigated. In  \cite{zhang2016control}, an optimal policy for autonomous vehicles is proposed, which considers both global service fairness and future costs. Yet, 
idle driving distance and real-time GPS information are not considered. In \cite{yang2004real,bertsimas2019online}, scheduling temporal-based methods are developed to reduce costs of idle cruising, however, the proposed algorithm does not utilize the real-time location information.
Minimizing the total customer waiting time by concurrently dispatching multiple taxis and allowing taxis to exchange their booking assignments is investigated in \cite{seow2010collaborative}.
Several studies in the literature, e.g., \cite{miao2016taxi,zhang2017taxi,oda2018movi,oda2018distributed}, have shown that learning from past taxi data is possible. Thus, taxi fleet management and minimizing both waiting-time for passengers and idle-time for drivers can be obtained. %In \cite{kheterpal2018flow}, an open source based on reinforcement learning for traffic control is developed, where traffic flow in a wide variety of traffic scenarios is improved. 
%A new set of benchmarks for mixed-autonomy traffic that expose several key
%aspects of traffic control is discussed in \cite{vinitsky2018benchmarks,kreidieh2018dissipating}. 
%However, the  settings, objective and formulations in our MHRS is different. %The authors in \cite{nam2017model,oda2018distributed} have
%used deep learning to solve traffic problems, such as travel mode choice prediction.
In \cite{oda2018movi}, authors used DQN for dynamic fleet management. Using realistic data, distributed DQN based approaches have shown to outperform the centralized solutions. This work is extended to accommodate the ride-sharing scenarios in \cite{alabbasi2019deeppool}. In this paper, we generalize these models and consider a multi-hop ride-sharing system where a rider can reach a destination via a number of transfers (hops) and at the same time one or more passengers can share a single vehicle.
With MHRS, the availability and range of the rideshare service increase, and also the total travelled vehicular miles can decrease.

%
%
% or do not address the problem of matching and dispatching holistically. In [], authors use DQN to solve the ride sharing problem but did not address the problem of multi hop ride sharing. 

\subsection{Our Contribution}

In this paper, we propose a distributed model-free multi-hop ride-sharing algorithm for (i) dispatching vehicles to locations where future demand is anticipated, and (ii) matching vehicles to customers.
We provide an optimized framework for vehicle dispatching that adopts deep  reinforcement learning to learn the optimal policies for each vehicle individually by interacting with the external environment. Our approach allows customers to have one (or more) hops along the way to the destination. 
We note that customers who accept to have hop(s) along their paths to destination may experience a slightly longer time than direct path. However, this incurred time could be compensated by giving incentives to the customers (e.g., lower payment rate) to opt for the carpooling with multi-hop. Our results show a small increase in the total trip times as compared to the rid-sharing scenario (see Figure \ref{hoping_Example} and its description for further details).

Different from the majority of existing model-based approaches, we do not need to accurately model the system (so we call it model-free), rather, we use deep reinforcement learning technique in which each vehicle solves its own double deep Q-network (DDQN) problem in a distributed manner without coordination with cars in its vicinity. By doing so, a significant reduction in the complexity is obtained. To the best of our knowledge,
this work is the first to cast the MHRS problem to a reinforcement learning problem. We aim to optimally dispatch the vehicles to different locations in a service area as to minimize a set of performance metrics including customers waiting time, extra travel time due to participating in MHRS and idle driving costs.  Using real-world dataset of taxi
trip records in New York City, our extensive simulation results show that, with the same overhead time as DeepPool \cite{alabbasi2019deeppool}, i.e., model-free algorithm for ride-sharing, passengers can complete their trips with $\sim 30\%$ lower costs. Further, MHRS can  achieve upto $\sim99\%$ accept rates, and thus minimizes the supply-demand mismatch. In addition, MHRS algorithm also helps  reduce both waiting time and (idle) cruising time by $\sim40\%$, with $\sim20\%$ better utilization rate of vehicles.

\if 0
We propose a novel model free approach to dispatching vehicles in multi-hop ride sharing. We show that with the same overhead time as ride sharing, passengers can complete their trips \char`\~30\% lower costs. We also show that multi-hop ride sharing achieves \char`\sim 99\% accept rates, thus minimizing the supply demand gap. Multi-hop ride sharing also helps in reduction of wait time as well as cruising time by\char`\~40\% and \char`\~20\% better utilization of the taxi fleet.

To the best of our knowledge, our work in multi hop ride sharing is the first to show how deep Q learning can be used to optimally dispatch vehicles in a ride hailing scenario and achieve better results than the existing approaches to ride sharing. 
\fi 
% \end{document}

\subsection{Organization}

This paper is organized as follows. Section II explains the problem and present an example for MHRS and distinguishes it from ridesharing problem. In addition, the used notations and performance metrics used in this paper are briefly explained.  
Section III explains the proposed MHRS framework and describes the main components in the system design. In addition, a distributed version of our algorithm is further presented in this section. Section IV presents the simulator design and the evaluation results. It also highlights the main findings out of this work. Section V concludes the paper and gives possible venues for promising directions and future research.

% \documentclass{article}
% \usepackage{graphicx}
% \graphicspath{ {./images/} }
% % if you need to pass options to natbib, use, e.g.:
% %     \PassOptionsToPackage{numbers, compress}{natbib}
% % before loading neurips_2019

% % ready for submission
% \usepackage{neurips_2019_ml4ad}

% \begin{document}
% % \begin{figure}
% % 	\begin{center}
% % 		\includegraphics[trim=0.1in 2.1502in 0.2in .0in, clip,width=0.48\textwidth]{figs/rideShareFig_v2}
% % 			\vspace{-.1in}
% % 		\caption{A schematic to illustrate the ride-sharing routing in a region graph,  consisting of 11 regions, $A$ to $K$. There are four ride requests and two vehicles. The locations of both customers and vehicles are shown in the figure above. Two different possible scenarios to serve the ride requests are shown in the figure and depicted by the dashed-red, dotted-green, and dashed-dotted blue lines. The destination of Rider 1 and Rider 4 is the Central Park, NY, while the destination of Rider 2 and Rider 3 is Times Square, NY.  	
% % 		}
% % 		\label{rideSharing_Example}
% % 		\vspace{-.2in}
% % 	\end{center}
% % \end{figure} 

\section{Problem Statement} \label{sec_probForm}

We consider a MHRS scenario where customers can complete their trips in multiple hops. We develop an algorithm that dispatches vehicles to either pick-up passengers from their pickup and/or hop locations or to serve potential customers in the dispatched zones. The vehicle location, capacity, next destination, and number of passengers, etc. can be tracked in real time, thanks to  mobile internet technologies. Without loss of generality, we consider the New York City area as our area of operation. The area is divided into multiple non-overlapping zones (or regions), each of which is $1$ square mile. This allows to discretize the area of operation and thus makes the action space (i.e, where to dispatch the vehicles) tractable. 

% We assume that the area is divided into several non-overlapping zones. The zone is assumed to be small (e.g., 1-sq mile grid). A zone comprises of several locations. We take the decision which zones the vehicles should go. If the number of zones is large, naturally, the number of decision variables will increase. Each zone has 2 dimensions corresponding to longitude and latitude. Note that ride hail services such as Uber, Lyft always provide the heat map the zone where the demand is high. 

\subsection{Multi-hop Example}
\begin{figure}
	\begin{center}
		\includegraphics[trim=0.1in .0in 0.1in .0in, clip,width=0.5\textwidth]{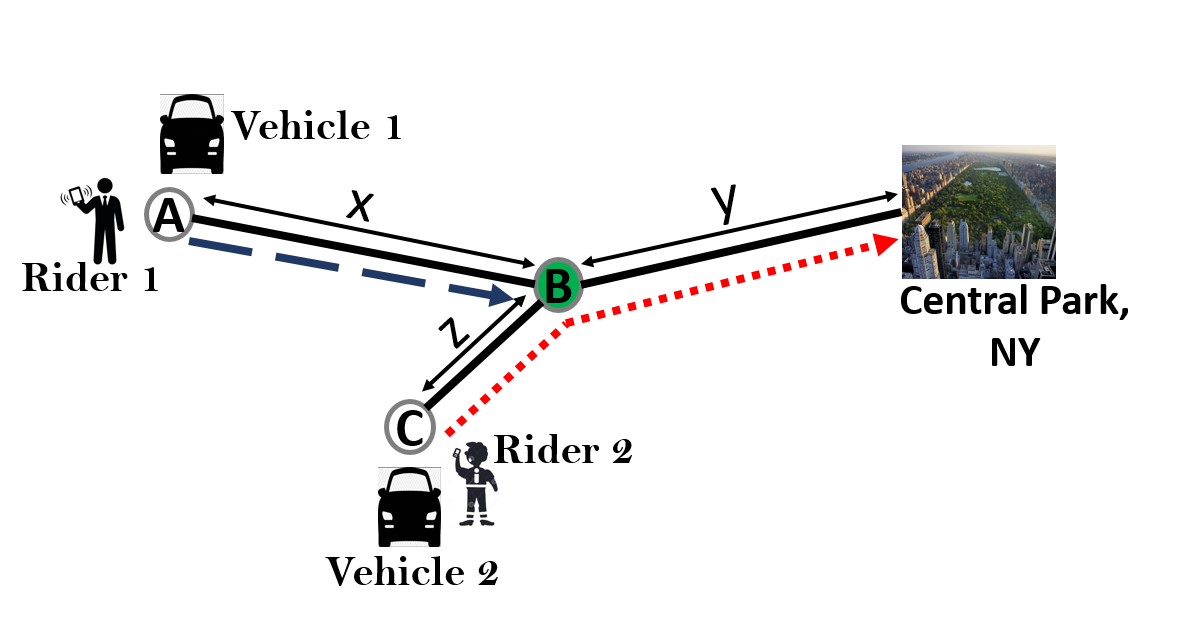}
%			\vspace{-.1in}
		\caption{A schematic illustrating the MHRS in a simple ride matching scenario. There are two vehicles (Vehicle $1$ and Vehicle $2$) and two ride requests (Ride $1$ and Ride $2$) at zones $A$ and $B$, respectively. Zone $B$ repesents at "hop" zone where one passenger changes/transfers to another vehicle in zone $B$.	
		}
		\label{hoping_Example}
%		\vspace{-.15in}
	\end{center}
\end{figure}

%{\it \textbf{Multi-hop Example}}:
 Consider a simple multi-hop ride-sharing scenario as depicted in Figure \ref{hoping_Example}. In this figure, both Rider $1$ and Rider $2$ want to go to the Central Park, NY. However, Rider $1$ and Rider $2$ are in different zones. Rider $1$ is in zone $A$, while Rider $2$ is in zone $C$. Since zone $B$ lies along the route of both Riders,  Zone $B$ can be a hop zone, where one vehicle (say Vehicle $1$) can drop its passenger and join the another vehicle (Vehicle $2$). Here, Rider $1$ initially gets into the nearest vehicle (Vehicle $1$) and Rider 2 gets into Vehicle $2$. Vehicle $1$ drops rider 1 at zone $B$. As Vehicle $2$ reaches the hop zone (zone $B$), it picks up Rider $1$ and drops both at their destination, i.e., Central Park. As shown in the figure, $x$ represents the distance between zone $A$ and zone $B$, $y$ represents the distance between zone $B$ and Central Park, and $z$ is the distance between zone $B$ and zone $C$. We define the effective distance of Vehicle $2$, $D_2$, which gives the effective distance traveled by Vehicle $2$, as follows:

\begin{equation}
    D_2 = \frac{z+y+y}{z+y}=1+\frac{y}{z+y}
    \label{eff_d}
\end{equation}

Thus, effective distance is formally defined as the ratio of total distance covered if no hoping and sharing was allowed to the total distance covered when hoping and sharing is allowed. We note that efficient packing of vehicles allows for less distance traveled by the vehicles in completing service of same number of requests. The multi hop scenario shown in Figure \ref{hoping_Example} reduces the traffic that goes from Zone $B$ to the Central Park. Further, MHRS allows vehicles to be free more frequently and thus can improve the accept rate of ride requests as in our example Vehicle $1$ is free after dropping Rider $1$ and can serve new customers accordingly.

% Thus we can define, effective distance $D_{effective }$ as:

% \begin{equation}\label{eq:effDist}
% \text{D_{effective}=\frac{\sum_{\ell=1}^{L}\sum_{n=1}^{N}D_{\ell,n}}{\sum_{\ell=1}^{L}(No hop distance)_{\ell}}}

% \end{equation}

% where $\ell$ denotes passengers, $n$ denotes vehicles and $D_{\ell,n}$ denotes distance traveled by passenger $\ell$ in vehicle $n$

% \begin{equation}
%  \[ x^n + y^n = z^n \]
% \end{equation}

\begin{figure*}
	\begin{center}
		\includegraphics[scale=1,width=0.75\textwidth]{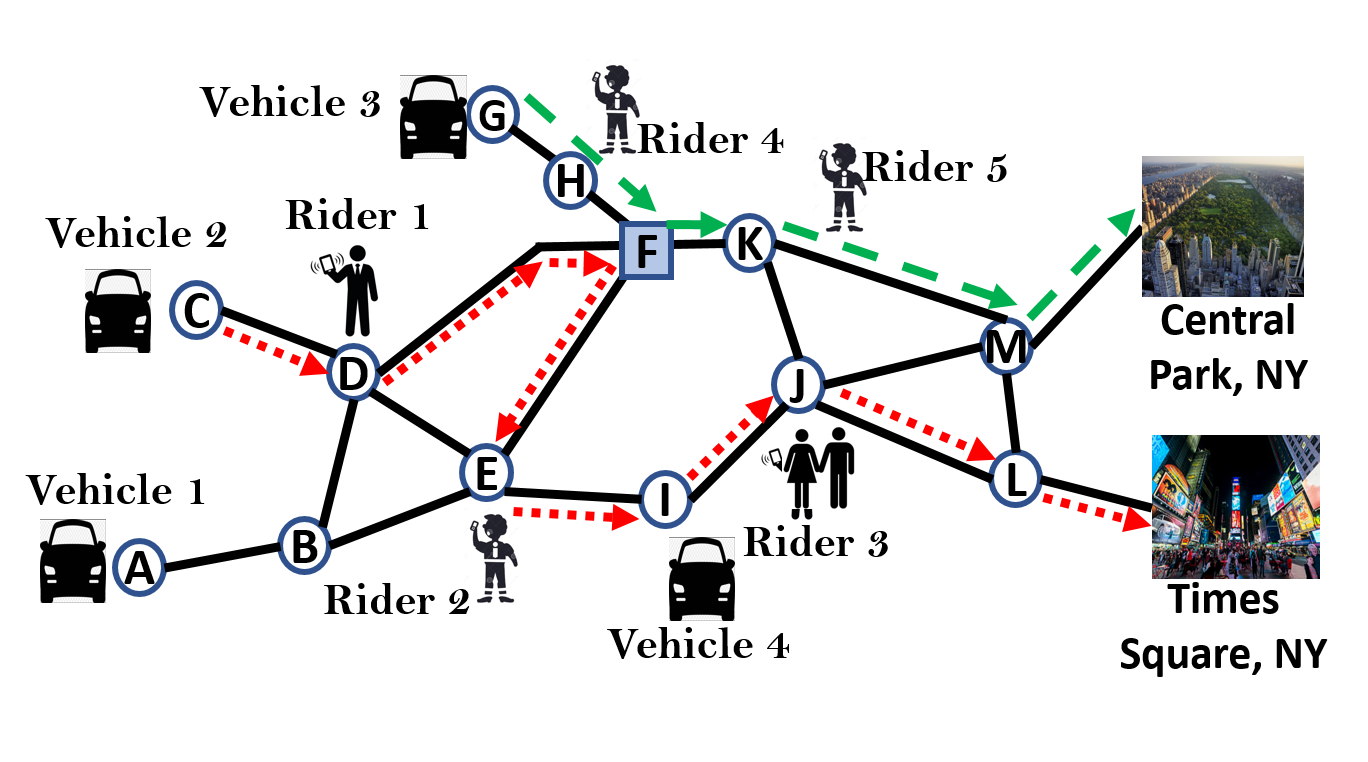}
		%		\vspace{-.2in}
		\caption{A schematic to illustrate the multi hop ride-sharing routing in a region graph,  consisting of 13 regions, $A$ to $M$. There are five ride requests and four vehicles. The locations of both customers and vehicles are shown in the figure above. The zone 'F' (colored in blue with a square shape) is the hop zone. The red dotted line shows the path of Vehicle 2 and green dotted line shows the path of Vehicle 3 in the multi-hop ride-sharing scenario. 	
		}
		\label{rideSharing_Example}
		%		\vspace{-.1in}
	\end{center}
\end{figure*}

%{\it  \textbf{MHRS, Ride-sharing, and No-ride-sharing Scenarios}}:

\subsection{MHRS, Ride-sharing, and No-ride-sharing Scenarios}
Figure \ref{rideSharing_Example} highlights the differences between different scenarios include:  ride-sharing with multi-hop, ride-sharing with no-hops, and the case where only one rider per vehicle is allowed (no ride-sharing case). The figure shows the locations of vehicles and the passengers at a given time. Riders 1, 4 and 5 want to go to the Central Park, NY, while Riders 2 and 3 want to go to the Times Square, NY. 
There can be a scenario where Vehicle 2 takes Rider 1,  and Vehicle 3 takes Rider 4 to the Central park. Further, vehicle 1 takes Rider 2 and Vehicle 4 takes Rider 3 to Times Square. In this case, no ride-sharing is involved and thus four vehicles are used to serve part of the demand. Since no more vehicles are available, Rider 5 request deemed rejected. We now consider a scenario where Vehicle 2 can serve Rider 1 (due to its proximity to Rider 1) and Vehicle 3 serves Riders 4 and 5. Vehicle 1 serves Rider 2 and Vehicle 4 serves Rider 3. In this case, ride-sharing allows efficient use of vehicles, thus no request is rejected. Still, there is a room for improvement through multi-hop ride-sharing. In such settings, Vehicle 2 picks-up Rider 1 from zone $D$ and drops Rider 1 at zone $F$ (i.e., a hop zone). Then, Vehicle 3 takes Rider 1, Rider 4 and Rider 5 to the Central Park (see the green dashed-line in Figure \ref{rideSharing_Example}). In addition, Vehicle 2 moves from zone $F$ to zone $E$ to pick up Rider 2 and then picks up Rider 3 a long its way to the Times square, NY  (the red dotted-line in Figure \ref{rideSharing_Example}). In this case,  only 2 vehicles are used to fulfill the demand. We note that under MHRS scenario, the non-used vehicles (Vehicle $1$ and Vehicle $4$) are free and able to serve other customers and thus improve the acceptance rate of ride-sharing systems.

From the proceeding discussion, we observe that MHRS leads to better utilization of vehicles, reduced travel cost due to sharing the cost by passengers, and reduced traffic congestion since less number of vehicles are used to serve the user's demand. Since traffic is reduced, pollution could be also mitigated. Moreover, since fewer number of vehicles are used, the incurred fuel cost is reduced, which results in less cost for customers.

\subsection{Model Parameters and Notation}

We use the following notations in our paper to define state, supply and demand. We use $i\in\{1,2,3,\ldots,M\}$ for the zones. The number of vehicles is taken as $N$. We optimize the multi hop system on $T$ time slots, where  one time slot has length $\Delta t$. The vehicles take decisions at each time slot $\tau=t_0,t_0+1,t_0+2,\cdots,t_0+T$, where $t_0$ is the current time slot.  Let the number of available vehicles at region $i$ at time slot $t$ be ${v}_{t,i}$. A vehicle is marked as {\em available} if at least one of its seats is vacant. A vehicle is marked unavailable if it is completely full or does not consider taking an extra passenger.
We can only dispatch those vehicles which are available. Vehicle $v$ has a maximum capacity of $C_v$ passengers. We take the number of requests at zone $i$ at time $t$ as ${d}_{t,i}$. $d_{t,\widetilde{t},i}$ is the number of vehicles that are not available at time $t$ but will drop-off a passenger(s) at zone $i$ and then become available at $\widetilde{t}$. We can estimate the $d_{t,\widetilde{t},i}$ using the estimated time of arrival (ETA) model \cite{alabbasi2019deeppool}.
We use $\boldsymbol{X}_t=\{\boldsymbol{x}_{t,1}, \boldsymbol{x}_{t,2},\ldots,\boldsymbol{x}_{t,N}\}$ to denote the vehicles status at time $t$. $\boldsymbol{x}_{t,k}$ is a vector which consists of the current zone of vehicle $k$, available vacant seats, the time at which a passenger is picked up, and destination zone of each passenger. Using this information, we can also predict the time slot at which the unavailable vehicle $v$ will be available,  $d_{t,\widetilde{t},i}$. Thus, for a set of dispatch actions at time $t$, we can predict the number of vehicles in each zone for $T$ time slots ahead, denoted by $\boldsymbol{V}_{t:t+T}$. We use ${hop}_{l,i,p}$ to denote the  passenger $l$ hopped down at zone $i$, and $p$ denotes the $p^{th}$ hop of the passenger.
We can improve upon our dispatching policies by estimating the demand in every zone through historical weekly/daily distribution of trips across zones \cite{wyld2005my}. 
We use $\boldsymbol{D}_{t:T}=\left( \overline{\boldsymbol{d}}_{t},\ldots,\overline{\boldsymbol{d}}_{t+T} \right) $ to denote the predicted future demand from time $t_0$ to time $t+T$ at each zone. Combining this data, the environment state at time $t$ can be written as $\boldsymbol{s}_t=\left(\boldsymbol{X}_t, \boldsymbol{V}_{t:t+T}, \boldsymbol{D}_{t:t+T} \right)$. Note that when a passenger's request is accepted, we will append the user's expected pick-up time, the source, and destination to $\boldsymbol{s}_t$.

\subsection{Objective}

The objective of this paper is to optimally dispatch the vehicles to various locations and achieve these motives:
a) ensure minimum waiting-time and dispatch/idle time for passengers and vehicles, respectively, b) minimize the gap between the supply and demand, c) minimize the number of used vehicles to serve the demand, d) minimize the number of hops (or transfers) for customers e) minimize the extra travel time for a customer for a given trip due to hops, and f) minimize the fuel consumption for each vehicle.

% Our objective is to efficiently dispatch the available fleet of vehicles to different locations in a given area in order to achieve the following goals: (i) satisfy the demand (or equivalently minimize the demand-supply mismatch), (ii) minimize the waiting time of the customers (time elapsed between the ride request and the pick up), as well as the dispatch time which is the time a vehicle takes to move to another zone to pick up new customer (may be, future customers), (iii) the extra travel time due to participating in ride-sharing, and (iv) the number of used vehicles/resources as to minimize the fuel consumption, congestion, and to better utilize the vehicles seats. 
%
%and the idle time (time in which a vehicle remains static in a zone) 
%(iii) the dispatch times- time in which a vehicle is not occupied, 

The metrics mentioned above will be explained in detail mathematically in the subsequent sections. We can associate different weights with these metrics. These weights help in deciding which metrics should be given more importance over others. For example, we can assign more weight to accept rate if we want to ensure high overall accept rate for the trips.

% We will explain these metrics mathematically in detail later (Section~\ref{sec:objective}). These metrics can be summed with different weights to give a combined objective. Further, the weights of these different components can be varied to reflect the importance of each one to the customers and fleet providers.

% Fig 2 shows a simple multi-hop scenario. Rider 1 has to go from A to Central Park, NY. Rider 2 has to go from C to Central Park. Point B is the hop zone. Rider 1 gets into Vehicle 1 and Rider 2 gets into Vehicle 2. Vehicle 1 drops rider 1 at point B. As Vehicle 2 reaches hop zone B, it picks up rider 1 and drops both at Central Park. Here, x is the distance between point A and point B, y is the distance between point B and Central Park and z is the distance between point B and point C.

% The effective distance, D, traveled by Vehicle 2 is this case is:

% \begin{equation}
%     \frac{z + y + y}{z + y}
% \end{equation}

For every time step, the agent takes an action, in our case the vehicles choose to be dispatched to the zones where future demand is anticipated. The agent (i.e., vehicle) receives a reward based on the action taken. Ideally, the agent takes the  actions that would maximize its expected discounted future reward, i.e.,
% Thus, at every time step $t$, the solver obtains a representation for the environment, $\boldsymbol{s}_t$, and a reward $\boldsymbol{r}_t$. Based on this information, it takes an action $\boldsymbol{a}_t$ to direct the idle vehicles to different locations such that the expected discounted future reward is maximized, i.e., 
\begin{equation}
\sum_{k=t}^{\infty}\eta^{k-t}\boldsymbol{r}_{k}(\boldsymbol{a}_{t},\boldsymbol{s}_{t}), \label{total_reward_all}
\end{equation}
where $\eta<1$ is a time discounting factor. In the context of this paper, the reward $\boldsymbol{r}_{k}(.)$ is defined as a weighted sum of different performance components that best capture the motive of our system as mentioned above. %Besides those metrics, the dispatch time for unoccupied vehicles - time to send a vehicle from one zone $i$ to another zone $j$- is an important component and should be minimized to avoid incurring additional running cost such as gasoline. Therefore, we try to not send a vehicle to far places as to reduce the cost. Moreover, we wish to minimize the additional travel time (i.e., extra time overhead) for participating in ride-sharing compared to the minimal travel time.
%
%, which are
%the total waiting
%
%dispatch time, the extra overhead time that would be incurred by pooling customers, number of rides that a vehicle $n$ can pick at time $t$, and the number of used vehicles (or alternatively the used resources/vehicles). The first component, dispatch time, reflects the time at which a vehicle is unoccupied and thus incurring running cost such as gasoline. We also try to not send a vehicle to a far place as to reduce the cost. 
%The second component gives the additional travel time for participating in ride-sharing compared to the minimal travel time.
% if  customer is assigned to a single vehicle for service. Thus, we wish to
Since the number of vehicles is limited,  it is likely that some rides may not be served, so we minimize the supply and demand gap. Since we want to motivate sharing and hopping but, at the same time, do not want the customer to have a bad experience, we minimize the overhead time and the number of hops per customer. In addition, we aim to minimize the wait time and idle cruising time to promote effective utilization of vehicles and smooth ride experience for the customer.
The agent takes an action $\boldsymbol{a}_t$ based on the environment state  $\boldsymbol{s}_t$ and receives the corresponding reward $\boldsymbol{r}_t$.

\section{Multi-Hop Ride-Sharing Framework Design} \label{MHRS_FW}

In this section, we present our framework to solve the MHRS problem. We propose a distributed model-free approach using DDQN, where the agent learns the best actions based on the reward it receives from the environment. It should be noted that the agent here refers to the vehicles, and our algorithm learns the optimal policy for each vehicle independently.

\subsection{Reward}
The first component of the reward is to minimize the gap between the supply and demand for the agents/vehicles. Let $\boldsymbol{v}_{t,i}$ denote the number of available vehicles at time $t$ at zone $i$. Supply demand difference within time $t$ at zone $i$ can be written as $(\boldsymbol{v}_{t,i}-\boldsymbol{\overline{d}}_{t,i})$, i.e., 
\begin{equation}\label{eq:mismatch}
\text{diff}^{(D)}_t=\sum_{i=1}^{M}\left(\overline{\boldsymbol{d}}_{t,i}-\boldsymbol{v}_{t,i}\right)^{+}%\label{diff_mismatch}
\end{equation}
where $(.)^+=\text{max(0,.)}$. The second component of the reward is to minimize the dispatch time for the vehicles. Vehicles can be dispatched in two cases, either to serve a new request or to move to  locations where a high demand is anticipated in the future. We note that only available vehicles can be dispatched. Dispatch time refers to the estimated travel time to go from the current zone to a particular zone, say zone $j$, i.e., $h^{n}_{t,j}$ if the vehicle $n$ is dispatched to zone $j$. Thus, for all available vehicles within time $t$, we wish to minimize the total dispatch time, $T^{(D)}_{t}$,
\begin{equation}\label{eq:cruise}
T^{(D)}_{t}=\sum_{n=1}^{N}\sum_{j=1}^{M}h^{n}_{t,j}u^{n}_{t,j} %\label{T_D}
\end{equation}
where $u^{n}_{t,j}=1$ only if vehicle $n$ is dispatched to zone $j$ at time $t$. We seek to minimize the dispatch time sine drivers are consuming fuel without gaining revenue.
Next, we want to minimize the extra travel time for every passenger (the third component).  For vehicle $n$, rider $\ell$, at time $t$, we need to minimize 
$\delta_{t,n,\ell}=t^{\prime}+t_{t,n,\ell}^{(a)}-t_{n,\ell}^{(m)}$, where $t_{t,n,\ell}^{(a)}$ is the updated time the vehicle will take to drop off passenger $\ell$ because of change in route and/or addition of a rider from the time $t$. $t_{t,n,\ell}^{(m)}$ is the  travel time that would have been taken if the passenger  $\ell$ would not have shared the vehicle with any other passenger. Also, $t^{\prime}$ is the time elapsed after the user $\ell$ has requested its ride. Therefore, we can write
\begin{equation}\label{eq:delt}
\boldsymbol{\Delta}_{t}=\sum_{n=1}^{N}\sum_{\ell=1}^{U_{n}}\delta_{t,n,\ell}
\end{equation}

To ensure that the passengers do not have to go through any inconvenience (e.g., many hops a long the way to destination) in order to complete their trips, we minimize the number of hops a passenger has to make to complete the trip. Recall that $hop_{\ell,i,p}$ denote the  passenger $\ell$ hopped down at zone $i$, and $p$ denotes the $p^{th}$ hop of the passenger $\ell$. Then,

\begin{equation}\label{eq:H}
{H}_{\ell}=\sum_{i=1}^{M}\sum_{p=1}^{P}hop_{\ell,i,p}
\end{equation}

Hence, the sum over all passengers $\ell$ at time $t$, gives the total number of hops of all customers at instant $t$, i.e., $\boldsymbol{H}_t=\sum_{\ell} H_{\ell}$.
Lastly, the number of vehicles at any time $\boldsymbol{e}_{t}$ needs to be optimized to better utilize the vehicles usage (i.e., available resources), this can be written as 

\begin{equation}
\boldsymbol{e}_{t}=\sum_{n=1}^{N}[\max\{\boldsymbol{e}_{t,n}-\boldsymbol{e}_{t-1,n},0\}] \label{activeVehicles}
\end{equation}
where $e_{t,n}$ shows whether a vehicle has been occupied. We want to minimize the number of vehicles in the fleet that change their status from being inactive to active. This would ensure that the vehicles in the fleet are utilized efficiently. This would help in reducing the idle cruising time of the vehicles, hence would lead to better vehicle utilization.

Now that we have obtained our main objective equations, we can define the accumulated reward equation, for all vehicles, at time $t$ as follows:

\begin{equation}
\overline{r}_t=-[\beta_1\text{diff}^{(D)}_{t}+\beta_2T^{(D)}_{t}+\beta_3\boldsymbol{\Delta}_{t}+\beta_4\boldsymbol{e}_{t}+\beta_5\boldsymbol{H}_{t}] \label{reward_cent}
\end{equation}

The minus sign indicates we want to minimize those terms, and  $\beta_1,\beta_2, \beta_3, \beta_4,\beta_5$ are weights and can be changed based on any specific objective we want to meet. For example, $\beta_i$ could be the cost per measure unit, e.g., cost per second. Moreover, we can increase $\beta_{4}$ if we decide our main goal to minimize the size of the fleet. Also, $\beta_i,\forall i$, can be vehicle dependent (i.e., $\beta_{i,n}, \forall i,n$) where each vehicle can choose its weights based on the drivers preference. We note that the reward is not an explicit function of the action. Through the model free approach, we learn the optimal relationship between action and reward by our algorithm.

We note that equation \eqref{reward_cent} gives the accumulated reward of all vehicles at time $t$ in a centralized manner. To solve the problem in a distributed fashion where each vehicle solves its own DDQN individually, we need to write the reward (and its related components) of every vehicle independent of other vehicles. In the next section, we provide the reward expression for each vehicle individually and show that by summing over all vehicles the total reward function is analogous to the objective function described in Section~\ref{MHRS_FW}.

\subsection{Distributed MHRS Framework}

We build a simulator to train and evaluate our framework. 
%Figure \ref{DP_arch} shows the basic blocks of DeepPool. The different blocks and the communications between them are depicted in the figure. 
We assume that a central unit is responsible for maintaining the states of all vehicles, e.g., current locations, destinations, occupancy state, etc. These states are updated in every time step based on the dispatching and assignments decisions. When needed, a vehicle $n$ communicates with the central unit to either request new information of other vehicles or update its own status. The goal of 
our distributed MHRS policy is to learn the optimal dispatch actions for each vehicle individually. Towards that goal, at every time slot $t$, vehicles decide $sequentially$ the next step (which zone to go) taking into consideration the locations of all other nearby vehicles in its vicinity. However, their current actions do not anticipate the future actions of other vehicles. We note that it is unlikely for two (or more) drivers to take actions at the same exact time since drivers know the location updates of other vehicles in real time, through the GPS for example. The advantage of MHRS policy stems from the fact that each vehicle can take its own decision without coordination with other vehicles. Next, we explain the state, reward and action.

\subsubsection{State}
The state variables capture the environment status and thus affect the reward feedback of different actions. The state at time $t$ is captured by a three tuple 
 $t$: $\left(\boldsymbol{X}_t, \boldsymbol{V}_{t:t+T}, \boldsymbol{D}_{t:t+T} \right)$.  These elements are combined in one vector denoted as $\boldsymbol{s}_t$. When a set of new ride requests are generated,  the MHRS engine updates its own data to keep track the environment status. The three-tuple state variables in $\boldsymbol{s}_t$ are pushed as an input to the neural network input layer  to make a certain decision. 

\subsubsection{Action}
We denote the action of vehicle $n$ by  $\boldsymbol{a}_{t,n}$. This action consists of two components --1) if the vehicle is partially filled it decides whether to serve the existing passengers or to accept new customer, and ii) if it decides to serve a new customer or the vehicle is totally empty, it decides the zone it should head (or hop) to at time slot $t$, which we denote by $u_{t,n,i}$. Here, $u_{t,n,i}=1$  if the vehicle decides to serve a new customer and/or head to zone $i$, otherwise it is $0$. Note that if vehicle $n$ is full it can not serve any additional customer. In contrast, if a vehicle decides to serve current customers, it opts the shortest optimal route for reaching the destinations of the users. 

\subsubsection{Reward} 
We now represent the reward fully, and the weights which we put towards each component of the reward function for each vehicle if it is not full. 
%First, note that  the reward is $0$ if the vehicle decides to serve the existing passengers in it (if, any). We, thus, turn our focus on the scenario where the vehicle decides to serve a new user and it is willing to take a  detour at time $t$. 
The reward  $r_{t,n}$ for vehicle $n$ at time slot $t$ in this case is given by
\begin{align}
r_{t,n} & =r(\boldsymbol{s}_{t,n},\boldsymbol{a}_{t,n})\nonumber\\
& =\beta_{1}b_{t,n}-\beta_{2}c_{t,n}-\beta_{3}\sum_{\ell=1}^{U_{n}}\delta_{t,n,\ell}\nonumber\\
& -\beta_{4}\max\{e_{t,n}-e_{t-1,n},0\}-\beta_{5}\underset{\ell\in\mathcal{L}_{t,n}}{\text{max}}\{\boldsymbol{H}_{\ell}\}\label{reward_{v}eh_{n}}
\end{align}
where  $\beta_i$ is the weight for component $i$ in the reward expression. Note that $\boldsymbol{H}_{t,n}=\underset{\ell\in\mathcal{L}_{t,n}}{\text{max}}\{\boldsymbol{H}_{\ell}\}$. Further, $b_{t,n}$ denotes the number of customers served by vehicle $n$ at time $t$ while $c_{t,n}$ denotes the time taken by vehicle $n$ to hop or take a detour to pick up extra customers if the vehicle has available seats. $\delta_{t,n,\ell}$ denotes the additional time vehicle $n$ takes because of carpooling compared to the scenario if the vehicle would have served customer $\ell$ solely without any carpooling and/or hoping. Note that while taking the decision, the vehicle $n$ exactly knows the destination of the passenger. Note that when a user is added, the route is updated for dropping the passengers. $U_n$ is the total number of chosen customers for pooling at vehicle $n$ till time $t$. $\mathcal{L}_{t,n}$ contains the set of all customers assigned to vehicle $n$ at instant $t$. Note that both $U_n$ and $\mathcal{L}_{t,n}$ are not known a priori and will be adapted dynamically in the MHRS policy. These two quantities will also vary as the passengers being picked or dropped by the vehicle $n$. 
% In Figure \ref{rideSharing_Example}, $U_4$ will be equal to $4$ if vehicle $2$ decided to serve all ride requests.
%The first term in \eqref{reward_veh_n} gives the number of rides that vehicle $n$ can pick up at time $t$, $u_{t,n}$, analogous to that in \eqref{diff_mismatch}. The second component gives the dispatch time or the travel time from one zone to another in order to pick up or look for a customer. Further, the third term corresponds to the extra travel time that a customer may be incurred when participating in ride-sharing. This term is formally defined in equation \eqref{T_D}.
%We can formally define $\delta$ as the difference between the travel time of the customer and the minimal travel time (which can be obtained by adopting the shortest-path between two nodes).
The last term captures the vehicle status where $e_{t,n}$ is set to one if empty vehicle $n$ becomes occupied (even if by one passenger), however, if an already occupied vehicles takes a new customer it is $0$. The intuition behind this reward term is that if an already occupied vehicle serves a new user, the congestion cost and fuel cost will be less rather than an empty vehicle serves the user.  
Note that if we make both $\beta_3$ and $\beta_5$ very large, we resort to the scenario where there is no carpooling. Since high $\beta_3$ indicates that passengers will not prefer detours to serve another passenger. Thus, the setting becomes similar to the one in \cite{oda2018movi}. Moreover, by setting $\beta_5$ to a very large value, our MHRS algorithm will resort to that of ride-sharing with no passenger transfer capability, which is similar to the algorithm presented in \cite{alabbasi2019deeppool}.

\subsection{Selection of Hop Zones}

Hop zones are the zones where customers hop down from a vehicle and wait for the next vehicle to complete the rest of their trips. We start by dividing the city into 212 x 219  square grids. To run our DDQN algorithm, we further combine grids to form a new map of 41 x 43 square grids, preventing our action space from exploding. Every third intersection of zones in the horizontal and vertical direction is considered a hop zone. Since we want to make sure that the hop zones are in the busy area of the city and are not too closely placed, we consider only the zones with at least 10 ride requests in the day. Through this method, we obtain 148 hop zones in the whole New York city.

%The proposed approach is shown in Algorithm \ref{MHRS_alg}. As shown in the algorithm, we first construct the state vector $\boldsymbol{\varOmega}_{t,n}$, using the ride requests and available vehicles, which are extracted from real records (Step 2). Next, for every available (not fully occupied) vehicle $n$, we select the action that maximizes its own reward, i.e., taking the {\it argmax} of the DQN-network output. The output of the DQN is the Q-values corresponding to every possible movement, while the input is the environment status governed by the vector $\boldsymbol{s}_t$ (Steps 2 and 3).
%Then, we update the solution by adding the tuple $(n,f_{n,t})$ to the set of dispatch solution (Step 2). Finally, $\boldsymbol{X}_{t:t+T}$ is updated based on the chosen action (line 7).  
%We note that the updates is performed sequentially so that vehicles can take other vehicles' actions into consideration when making their own decisions. Nonetheless, each vehicle does not anticipate the future actions of other vehicles, thus limiting the coordination among them. 

\subsection{MHRS Algorithm Description}
In this section, we explain our proposed algorithm for MHRS in detail. The full algorithm is shown in Algorithm \ref{MHRS_alg}. We first construct the state vector $\boldsymbol{\varOmega}_{t,n}$, using the ride requests and available vehicles, which are extracted from real records (line 1). Then, we initialize the number of vehicles and generate some ride requests in each time step based on the real record in the dataset (see lines 2 to 5). 
Next, 
the agent determines the actions $\boldsymbol{a}_t$ using our MHRS algorithm and matches the vehicle to the appropriate ride request (or requests), lines 6 to 14. 
Every vehicle $n$ selects the action that maximizes its own reward, i.e., taking the {\it argmax} of the DDQN-network output. The output of the DDQN are the Q-values corresponding to every possible movement, while the input is the environment status governed by the vector $\boldsymbol{s}_t$. Note that each vehicle updates its status at the agent based on the chosen action (line 15). 
We note that the updates are performed sequentially so that vehicles can take other vehicles' actions into account when making their own decisions. Nonetheless, each vehicle does not anticipate the future actions of other vehicles, thus limiting the coordination among them. 
Then, the vehicles traverse to the dispatched locations using the shortest path in the road network graph. The dispatched vehicles travel to target locations within the time estimated by our model (Section IV-B), see lines 18 to 23. The extra travel time $\delta_{t,n}$ and hop counter $H_{t,n}$ are updated as needed (lines 19-25).

% \begin{algorithm}
% 	\caption{Multi Hop Algorithm for ride-sharing using DQN\label{alg:MutiHop}}
% 	\begin{enumerate}
% 		\item \textbf{Input:} $\boldsymbol{X}_t, \boldsymbol{V}_{t:t+T}, \boldsymbol{D}_{t:t+T}$;
% 		%		\item $\quad$//\textit{\small{}steps}{\small \par}
% 		\item $\quad$\textbf{Step 1}: Construct a state vector  $\boldsymbol{\varOmega}_{t,n}=\left(\boldsymbol{X}_t, \boldsymbol{V}_{t:t+T}, \boldsymbol{D}_{t:t+T} \right)$;
		
% 		\item $\quad$\textbf{Step 2}: Choose ${a}_{t,n}$ such that: 
		
% 		${a}_{t,n}=\underset{a}{\text{arg\,max}}Q(\boldsymbol{\varOmega}_{t,n},a;\theta)$;
		
% 		\item $\quad$\textbf{Step 3}: Get the destination zone $f_{t,n}$ for vehicle $n$ based on action $a_{t,n}$;

% 		\item $\quad$\textbf{Step 4}: Update the dispatch solution by adding $(n,f_{t,n})$
		
% 		%		\item $\quad$\textbf{Step 4}: Update the dispatch solution by adding $(n,f_{t,n})$
% 		\item \textbf{endfor}
% 		\item \textbf{output:} Dispatch Solution
% 	\end{enumerate}
% \end{algorithm}

\begin{algorithm}[t]
\caption{ MHRS Algorithm}
\label{MHRS_alg}
\begin{algorithmic}[1]
\State Form a state vector  $\boldsymbol{\varOmega}_{t,n}=\left(\boldsymbol{X}_t, \boldsymbol{V}_{t:t+T}, \boldsymbol{D}_{t:t+T} \right)$.
\State Initialize vehicles states $\boldsymbol{X}_0$ as location of first $N$ requests.
\For{$t \in$ {\it Total time}}
\State Get all ride requests at time $t$
\State Group all the requests assigned to the same hop zone
\For{Each request group in time slot $t$}
    \State \textbf{Match} a vehicle $n$ to serve the requests
    \State \textbf{Update} the vehicle capacity
    \State \textbf{Update} the pickup and destination for hop trips 
    \State \textbf{Calculate} wait time and the number of passengers \State without any hops
    \State \textbf{Calculate} the dispatch time using ETA model % \ref{RN_ETA}.
    \State \textbf{Update} the state vector $\boldsymbol{\varOmega}_{t,n}$.
    \EndFor
    
% \For{Each ride request(in hop zone) in time slot $t$}
%     \State \textbf{Choose} a vehicle $n$ to serve the hop requests
%     \State \textbf{Update} the vehicle capacity
%     \State \textbf{Calculate} the dispatch time using ETA model % \ref{RN_ETA}.
%     \State \textbf{Update} the state vector $\boldsymbol{\varOmega}_{t,n}$.
%     \EndFor
    \State \textbf{Send} the state vector $\boldsymbol{\varOmega}_{t}$ to the Q network.
    \State \textbf{Get} the best dispatch action $\boldsymbol{a}_t$ for each agent from \State the network.    
\For{ all the available vehicles $n \in \boldsymbol{a}_t$ }    
     \State \textbf{Update} $\boldsymbol{H}_{t,n}$, if needed, and update the current \State location of vehicle
     \State \textbf{Find} the shortest path to every dispatch location \State  for every vehicle $n$.
     \State \textbf{Estimate} the travel time using the ETA model
     \State \textbf{Update} $\delta_{t,n}$, if needed, and generate the trajectory \State  of vehicle $n$.
 \EndFor  
  \State \textbf{Update} the state vector $\boldsymbol{\varOmega}_{t+1}$.     
%    =\State $\left(\boldsymbol{X}_t, \boldsymbol{V}_{t:t+T}, \boldsymbol{D}_{t:t+T} \right)$. {$f(\overrightarrow{X}_i)$}
%    \If{$f(\overrightarrow{X}_i)<f(\overrightarrow{P}_i)$}
%       \State $\overrightarrow{P}_i \gets \overrightarrow{X}_i$
%    \EndIf
%    \If{$f(\overrightarrow{X}_i)<f(\overrightarrow{P}_g)$}
%       \State $\overrightarrow{P}_g \gets \overrightarrow{X}_i$
%    \EndIf
\EndFor
\end{algorithmic}

\end{algorithm} 

%\vspace{-.25in}

% \input{alg}
%  \end{document}
% \documentclass{article}
% % if you need to pass options to natbib, use, e.g.:
% %     \PassOptionsToPackage{numbers, compress}{natbib}
% % before loading neurips_2019

% % ready for submission
% \usepackage{neurips_2019_ml4ad}

% % to compile a preprint version, e.g., for submission to arXiv, add add the
% % [preprint] option:
% % \usepackage[preprint]{neurips_2019_ml4ad}

% % to compile a camera-ready version, add the [final] option, e.g.:
% % \usepackage[final]{neurips_2019_ml4ad}

% % to avoid loading the natbib package, add option nonatbib:
% % \usepackage[nonatbib]{neurips_2019_ml4ad}

% \usepackage[utf8]{inputenc} % allow utf-8 input
% \usepackage[T1]{fontenc}    % use 8-bit T1 fonts
% \usepackage{hyperref}       % hyperlinks
% \usepackage{url}            % simple URL typesetting
% \usepackage{booktabs}       % professional-quality tables
% \usepackage{amsfonts}       % blackboard math symbols
% \usepackage{nicefrac}       % compact symbols for 1/2, etc.
% \usepackage{microtype}      % microtypography
% \usepackage{graphicx}
% \graphicspath{ {./images/} }
% \begin{document}

\section{Simulator Setup and Evaluation}

\subsection{Setup and Initialization}
Through extensive simulations, we show how MHRS algorithm works in real time scenario as compared to ride-sharing \cite{alabbasi2019deeppool} and no ride-sharing \cite{oda2018movi} scenarios. We trained our algorithm using real world taxi trips in New York City for May 2016. Then, we tested our algorithm using the trip records of the first two weeks of June 2016 data.
We divide the New York city into grids of size $41\times 43$, each grid has a an area of 800 x 800 m$^{2}$.

To initialize the environment, we run the simulation for $20$ minutes without dispatching the vehicles. Further, we set the number of vehicles to $N=5000$. The initial locations of these vehicles correspond to the  
first $5000$ ride requests. We set the maximum horizon to $T=30$ step and $\Delta t =1$ minute. We breakdown the reward and investigate the performance of the different policies.
Further, unless otherwise stated, we set $\beta_1=5$, $\beta_{2}=1$, $\beta_{3}=3.5$, $\beta_{4}=0.05$, and $\beta_{5}=2$. 

Recall that the reward function is composed of five different metrics, which we seek to minimize: (i) mismatch between supply and demand, (ii) dispatch time for vehicles, (iii) extra travel time because of ride-sharing and/or detouring for hops, (iv) number of used vehicles (resources), and finally (v) number of hops per customer. We note that the supply-demand mismatch is reflected in our simulation through the accept rate metric. This metric is computed as the number of ride accepts divided by the total number of requests per day. Recall that a request is deemed rejected if there is no vehicle in a range of $5km^2$, or no vehicle is available to serve the request. Also, the metric of idle time represents the time at which a vehicle is not occupied while still incurring gasoline cost and not gaining revenue. In addition,  the waiting time is composed of two sub-components: the time from the ride request till the time at which the vehicle arrives to pick up the customer and the incurred wait time in hop zone(s), if any. Below, we explain our models to estimate the time of arrival, demand and our proposed DDQN architecture.

\subsection{Estimated Time of Arrival (ETA) and Demand Prediction }

\begin{figure}[htbp]
	\centering
	\includegraphics[trim=0.4in 0.3in 0.6in 0in, clip,width=0.5\textwidth]{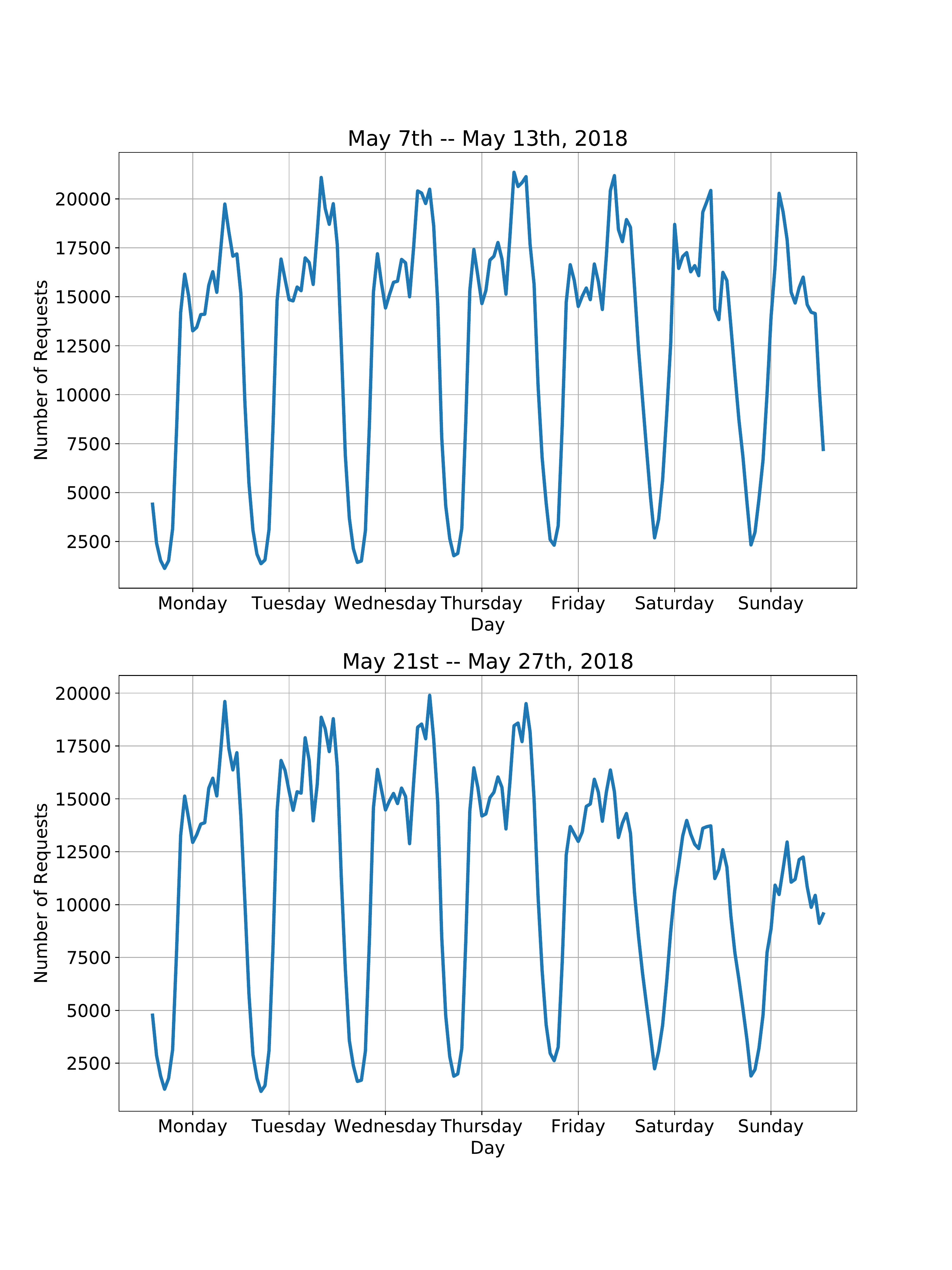}
		\caption{Demand pattern for two weeks in May 2018.}
		\label{demand}
\end{figure}

We use the New York City taxi data set \cite{manh_dataSet} to build the ETA and demand prediction models.
In the ETA model, we want to predict the expected travel time between two zones (two pairs of latitudes and longitudes). We split our data into 70\% train and 30\% test. We use day of week, latitude, longitude and time of days as our predictor variables and use random forest to predict the ETA. We obtain a root mean squared error (RMSE) of 3.4 on the test data.

\begin{figure}[h]
	\centering
	\includegraphics[width=0.5\textwidth,scale=1]
	{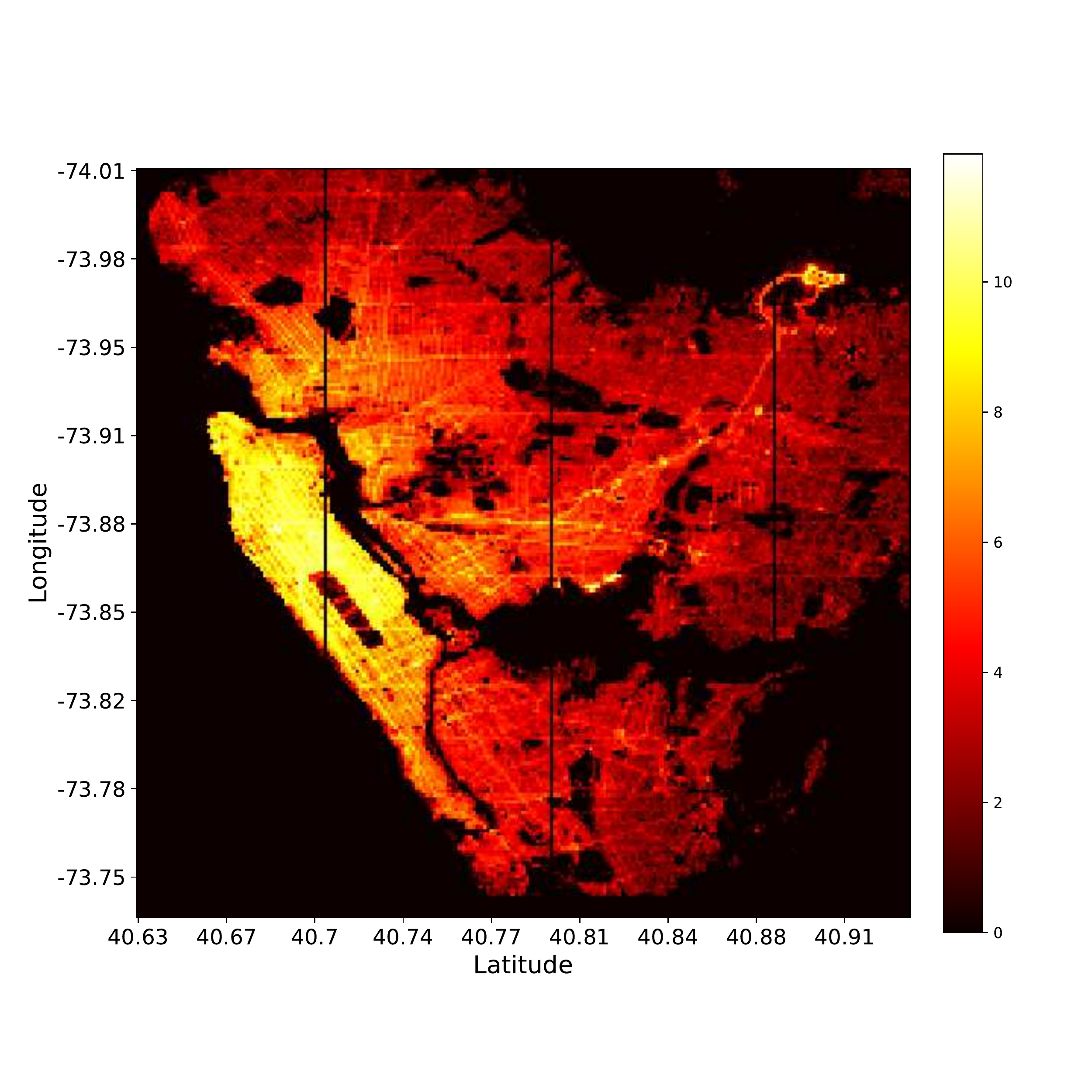}
		\caption{Demand Heat Map in New York City}
		\label{heat_map}
\end{figure}

For the demand prediction model, we predict the expected number of requests in a given zone for $15$ and $30$ minutes ahead. 
We plot in Figure \ref{demand} the weekly number of ride requests for two different weeks: May 7th-May 14th and May 21st-May 27th. We observe that both two weeks data exhibits the same daily behavior. In particular, both datasets experience daily periodicity with lower demand on the weekend, and thus the demand has a daily/weekly pattern that can be predicted. In Figure \ref{heat_map} we show the demand pattern in the form of a heat map. The New York city is plotted in the form of a rectangular grid of size 212 x 219. The brightest areas on the heat map, yellow zones, belong to the Manhattan area, where the trip demands are very high throughout the day.

We use convolutional neural networks for our demand predictions. The input to the network is 2 planes of 212 x 219. Each grid in the planes represents the ride request in that zone for the previous 2 time steps. In the network, we have 16 of 5x5 filters, 32 of 3x3 filters and finally one 1x1 filter. The output is a plane of 212x219, with each grid in the plane representing the demand in that zone in the next 30 minutes. The output of the demand prediction model will be used in the state in our model free algorithm. It should be further noted that we are not explicitly learning the transition probabilities from one state to another, which makes our approach model free. 

\subsection{Deep Q-Network Architecture}

For the DDQN, we divide the New York City area map to obtain 41 x 43 grids. We do so to limit the action space of vehicles  to $7$ grids vertically and horizontally. Thus the action space for each vehicle is 15 x 15 grid around its present location. The input to the deep Q network is the state of vehicles, supply and demand. Specifically, it consists of predicted requests in the next 15 minutes obtained from demand model, current location of each vehicle, future location of each vehicle in the next 15, and 30 minutes. The network architecture consists of 16 convolution layers of 5 x 5, 32 convolution layer of 3 x 3, 64 convolution layer of 3 x 3, 16 convolution layer of 1 x 1. All convolution layers have a rectifier non linearity activation. The output of the deep Q network is 15x 15 Q values, where each value corresponds to the reward a vehicle could get if dispatched to that particular zone.

Reinforcement learning becomes unstable for non linear approximations, hence we use experience replay to overcome this issue. We define a new parameter, $\alpha$ to address this issue. We memorize the experiences and randomly sample from these experiences to avoid correlation between immediate actions and experiences. The value of $\alpha$ is an indication of how frequently the target Q values are updated, indicating how much the agents are learning from their past experiences. In our case, we have taken $\alpha$ as 0.9
% We increase the value of $\alpha$ from 0.1 to 1, meaning only 10\% of the vehicles take action in the beginning. 
%A detailed step-by-step description for our MHRS algorithm is as shown in Algorithm \ref{MHRS_alg}.

\subsection{Evaluation and Results}
We compare the results of our MHRS algorithm with two scenarios: no ride-sharing (No-RS) and ride-sharing (RS). No-RS means the vehicles are dispatched using DDQN policies but only one rider can occupy a vehicle at any given time. This policy is akin to that in \cite{oda2018movi}, which we refer in our simulation comparisons as "MOVI". For the RS case, in addition to DDQN dispatch policies, more than one passenger can be assigned to one vehicle but no multi-hop is performed. This policy is proposed in \cite{alabbasi2019deeppool}, so-called DeepPool. In our MHRS policy, ridesharing is allowed and passengers could complete their trips in more than one hop.

Unless otherwise stated, we set up the simulator with 5000 vehicles. The time step was taken as one minute and the agents were trained on data of one month duration and tested on data of two weeks (first two weeks of June 2016). The initial location of the vehicles are corresponding to the locations of the first 5000 trips.

% \begin{figure}[h]

% \includegraphics[scale=0.5]{{"images/accept rate num hops"}.png}

% \caption{Accept Rate for average number of hops}
% \label{Fig. 1}
% \end{figure}

%\input{test.tex}

\begin{figure*}[h]
\centering
\hspace{-0.5cm}
        %\hspace{-0.5in}
		\begin{minipage}[b]{0.32\linewidth}
		% This file was created by tikzplotlib v0.8.4.
\begin{tikzpicture}[scale=0.75]

\begin{axis}[
legend cell align={left},
legend style={draw=white!80.0!black},
tick align=outside,
tick pos=left,
x grid style={white!69.01960784313725!black},
xlabel={No of Hops},
xmajorgrids,
xmin=1.34207615770323, xmax=1.47998988128987,
xtick style={color=black},
y grid style={white!69.01960784313725!black},
ylabel={Accept Rate},
ymajorgrids,
ymin=0.916413910437907, ymax=0.995307880803951,
ytick style={color=black}
]
\addplot [very thick, green!50.19607843137255!black, mark=*, mark size=4, mark options={solid}]
table {%
1.34834496332081 0.991721791241858
1.36386536652126 0.985896567776884
1.3687000907441 0.983335379669833
1.38068179641865 0.979537742854189
1.41396643155923 0.96799252625909
1.41408837626154 0.968231437989672
1.47372107567229 0.950165872314534
};
\addlegendentry{MHRS}
\addplot [very thick, red, dashed, mark=triangle*, mark size=4, mark options={solid,rotate=180}]
table {%
1.34834496332081 0.95
1.36386536652126 0.95
1.3687000907441 0.95
1.38068179641865 0.95
1.41396643155923 0.95
1.41408837626154 0.95
1.47372107567229 0.95
};
\addlegendentry{DeepPool}
\addplot [very thick, blue, dashed, mark=square*, mark size=4, mark options={solid}]
table {%
1.34834496332081 0.92
1.36386536652126 0.92
1.3687000907441 0.92
1.38068179641865 0.92
1.41396643155923 0.92
1.41408837626154 0.92
1.47372107567229 0.92
};
\addlegendentry{MOVI}
\end{axis}

\end{tikzpicture}
		\vspace{-.3in}
% \includegraphics[scale=1,width=\textwidth]{{"images/accept rate num hops v3"}.png}
%\vspace{-.3in}
\caption{Accept rate versus the average number of hops.}
\label{accept_vs_hop}
%			\vspace{2ex}
		\end{minipage}	
	\hspace{.35cm}
		\begin{minipage}[b]{0.32\linewidth}
		% This file was created by tikzplotlib v0.8.4.
\begin{tikzpicture}[scale=0.75]

\begin{axis}[
legend cell align={left},
legend style={at={(0.03,0.97)}, anchor=north west, draw=white!80.0!black},
tick align=outside,
tick pos=left,
x grid style={white!69.01960784313725!black},
xlabel={No of Hops},
xmajorgrids,
xmin=1.34393808601652, xmax=1.44088938671084,
xtick style={color=black},
y grid style={white!69.01960784313725!black},
ylabel={Effective Distance Ratio},
ymajorgrids,
ymin=0.979264023567787, ymax=1.43545550507648,
ytick style={color=black}
]
\addplot [very thick, green!50.19607843137255!black, mark=*, mark size=4, mark options={solid}]
table {%
1.34834496332081 1.20249286952973
1.36629533735123 1.30830525707081
1.38068179641865 1.35377555373121
1.41088480317118 1.39929156338145
1.43648250940655 1.41471952864427
};
\addlegendentry{MHRS}
\addplot [very thick, red, dashed, mark=triangle*, mark size=4, mark options={solid,rotate=180}]
table {%
1.34834496332081 1.15
1.36629533735123 1.15
1.38068179641865 1.15
1.41088480317118 1.15
1.43648250940655 1.15
};
\addlegendentry{DeepPool}
\addplot [very thick, blue, dashed, mark=square*, mark size=4, mark options={solid}]
table {%
1.34834496332081 1
1.36629533735123 1
1.38068179641865 1
1.41088480317118 1
1.43648250940655 1
};
\addlegendentry{MOVI}
\end{axis}

\end{tikzpicture}
			\vspace{-.3in}
			\caption{Effective distance ratio versus number of hops.}
			\label{Effdist_vs_hops}
%			\vspace{2ex}
		\end{minipage}
	\hspace{.28cm}
		\begin{minipage}[b]{0.32\linewidth}
		% This file was created by tikzplotlib v0.8.4.
\begin{tikzpicture}[scale=0.75]

\begin{axis}[
legend cell align={left},
legend style={at={(0.03,0.03)}, anchor=south west, draw=white!80.0!black},
tick align=outside,
tick pos=left,
x grid style={white!69.01960784313725!black},
xlabel={Accept Rate},
xmajorgrids,
xmin=0.939805171611156, xmax=0.988091396165729,
xtick style={color=black},
y grid style={white!69.01960784313725!black},
ylabel={Idle Time (min)},
ymajorgrids,
ymin=1.53621013481654, ymax=5.68875189834207,
ytick style={color=black}
]
\addplot [very thick, green!50.19607843137255!black, mark=*, mark size=4, mark options={solid}]
table {%
0.943179818597732 4.09449870342675
0.964075652449547 3.29559727166491
0.979537742854189 2.63187893839617
0.985896567776884 1.72496203315861
};
\addlegendentry{MHRS}
\addplot [very thick, red, dashed, mark=triangle*, mark size=4, mark options={solid,rotate=180}]
table {%
0.942 5.151342685634
0.96 5.516453528352
0.97 5.45246354287473
0.98 5.42654537828532
};
\addlegendentry{DeepPool}
\addplot [very thick, blue, dashed, mark=square*, mark size=4, mark options={solid}]
table {%
0.942 5.11652421
0.96 5.41374736383
0.97 5.35236675846
0.98 5.327635352833
};
\addlegendentry{MOVI}
\end{axis}

\end{tikzpicture}
			\vspace{-.3in}
			\caption{Average idle time per request for different accept rate.}
			\label{idle_vs_accept}
		\end{minipage}
%		\hspace{.05in}
%		\begin{minipage}[b]{0.32\linewidth}
%			\includegraphics[scale=1,width=\textwidth]{{"images/wait time accept rate v2"}.png}
%			\vspace{-.3in}
%			\caption{Average wait time per request for different accept rate.}
%			\label{wait_vs_accept}
%		\end{minipage}
%		\hspace{.05in}
%		\begin{minipage}[b]{0.32\linewidth}
%		\includegraphics[scale=1,width=\textwidth]{{"images/No of Vehicles accept rate v2"}.png}
%		\vspace{-.3in}
%		\caption{Average vehicles used for different accept rate}
%		\label{veh_vs_accept}
%	\end{minipage}	
% 	\hspace{.01in}
% 	\begin{minipage}[b]{0.32\linewidth}
% 		\includegraphics[trim=0.25in 0.02in 4.0in .0in, clip,width=\textwidth]{figs2/idleTime_vs_extraOverhead_v1}
% 		\vspace{-.3in}
% 		\caption{Average idle time per vehicle for different overhead values $\delta_t$.}
% 		\label{idleTime_vs_overhead}
% 	\end{minipage}
\end{figure*}

\begin{figure*}[t]
	\centering
	\hspace{-0.42cm}
	\begin{minipage}[b]{0.32\linewidth}
	% This file was created by tikzplotlib v0.8.4.
\begin{tikzpicture}[scale=0.75]

\begin{axis}[
legend cell align={left},
legend style={draw=white!80.0!black},
tick align=outside,
tick pos=left,
x grid style={white!69.01960784313725!black},
xlabel={Accept Rate},
xmajorgrids,
xmin=0.947913910437907, xmax=0.993807880803951,
xtick style={color=black},
y grid style={white!69.01960784313725!black},
ylabel={Wait Time (min)},
ymajorgrids,
ymin=0.531479978572822, ymax=5.73659619149653,
ytick style={color=black}
]
\addplot [very thick, green!50.19607843137255!black, mark=*, mark size=4, mark options={solid}]
table {%
0.95 1.49252765227841
0.97 1.39589523272865
0.98 1.01875393442224
0.99 0.768076170069355
};
\addlegendentry{MHRS}
\addplot [very thick, red, dashed, mark=triangle*, mark size=4, mark options={solid,rotate=180}]
table {%
0.95 4.1437685735324
0.97 3.2127589365473
0.98 3.1257363493628
0.99 2.9117894753227
};
\addlegendentry{DeepPool}
\addplot [very thick, blue, dashed, mark=square*, mark size=4, mark options={solid}]
table {%
0.95 5.51289674156394
0.97 3.75238096736538
0.98 3.25218995683393
0.99 3.11199755384932
};
\addlegendentry{MOVI}
\end{axis}

\end{tikzpicture}
		\vspace{-.3in}
		\caption{Average wait time per request for different accept rate.}
		\label{wait_vs_accept}
	\end{minipage}
	\hspace{.35cm}
	\begin{minipage}[b]{0.32\linewidth}
	% This file was created by tikzplotlib v0.8.4.
\begin{tikzpicture}[scale=0.75]

\begin{axis}[
legend cell align={left},
legend style={draw=white!80.0!black},
tick align=outside,
tick pos=left,
x grid style={white!69.01960784313725!black},
xlabel={Accept Rate},
xmajorgrids,
xmin=0.947913910437907, xmax=0.993807880803951,
xtick style={color=black},
y grid style={white!69.01960784313725!black},
ylabel={Wait Time (min)},
ymajorgrids,
ymin=0.531479978572822, ymax=5.73659619149653,
ytick style={color=black}
]
\addplot [very thick, green!50.19607843137255!black, mark=*, mark size=4, mark options={solid}]
table {%
0.95 1.49252765227841
0.97 1.39589523272865
0.98 1.01875393442224
0.99 0.768076170069355
};
\addlegendentry{MHRS}
\addplot [very thick, red, dashed, mark=triangle*, mark size=4, mark options={solid,rotate=180}]
table {%
0.95 4.1437685735324
0.97 3.2127589365473
0.98 3.1257363493628
0.99 2.9117894753227
};
\addlegendentry{DeepPool}
\addplot [very thick, blue, dashed, mark=square*, mark size=4, mark options={solid}]
table {%
0.95 5.51289674156394
0.97 3.75238096736538
0.98 3.25218995683393
0.99 3.11199755384932
};
\addlegendentry{MOVI}
\end{axis}

\end{tikzpicture}
		\vspace{-.3in}
		\caption{Average used vehicles for different accept rate.}
		\label{veh_vs_accept}
	\end{minipage}
	\hspace{.01in}
\begin{minipage}[b]{0.32\linewidth}
    % This file was created by tikzplotlib v0.8.4.
\begin{tikzpicture}[scale=0.75]

\begin{axis}[
legend cell align={left},
legend style={draw=white!80.0!black},
tick align=outside,
tick pos=left,
x grid style={white!69.01960784313725!black},
xlabel={No of Vehicles},
xmajorgrids,
xmin=4800, xmax=9200,
xtick style={color=black},
y grid style={white!69.01960784313725!black},
ylabel={Idle Time(min)},
ymajorgrids,
ymin=0, ymax=8,
ytick style={color=black}
]
\addplot [very thick, green!50.19607843137255!black, mark=*, mark size=4, mark options={solid}]
table {%
5000 0.60530493720226
6000 0.625311609046147
7000 1.44146914663023
8000 1.43540307079912
9000 1.18076399274018
};
\addlegendentry{MHRS}
\addplot [very thick, red, dashed, mark=triangle*, mark size=4, mark options={solid,rotate=180}]
table {%
5000 5.151452136784321
6000 5.523825316732145
7000 5.454287541396323
8000 5.411567432567497
9000 5.611754229563422
};
\addlegendentry{DeepPool}
\addplot [very thick, blue, dashed, mark=square*, mark size=4, mark options={solid}]
table {%
5000 4.132675344238482
6000 5.428744830924134
7000 5.351545833285532
8000 5.311265835246722
9000 5.643328653294632
};
\addlegendentry{MOVI}
\end{axis}

\end{tikzpicture}
	\vspace{-.3in}
	\caption{Average idle time  for different number of vehicles.}
	\label{veh_vs_idletime}
\end{minipage}
\end{figure*}

Figure \ref{accept_vs_hop} plots the average accept rate versus the number of hops. We change the parameter $\beta_{5}$ to show the effect of hops on the accept rate. The values of $\beta_{5}$ used are - $2$, $1.9$, $1.8$, $1.7$, $1.6$, $1.5$ and $1.4$. Note that if a passenger is dropped at a hop zone and not served in $\delta_{t,n,\ell}$, this passenger request is deemed rejected. Since DeepPool and MOVI are not function of the number of hops, both have fixed acceptance rates at $0.95$ and $0.92$, respectively.  We note that by $1.40$ hops, only $40\%$ of the total trips have $2$ hops.  
While the accept rate decreases as number of hops increases, our MHRS approach still performs the best as compared to both DeepPool and MOVI.  This decrease of accept rate is due to the possibility for our algorithm to fail serving some of the passengers at the hop zones. In addition, some passengers complete their trips in more than one hop, thus making some vehicles available at the hop zones. Hence, vehicles get free more often as compared to when they had to travel long distances in one trip.

Figure \ref{Effdist_vs_hops} shows the effective distance ratio of vehicles (defined in \eqref{eff_d}) versus the number of hops.
We change the parameter $\beta_{5}$ to show the effect of hops on the effective distance. The values of $\beta_{5}$ used are - $2$, $1.9$, $1.8$, $1.7$ and $1.6$. Increasing effective distance with number of hops suggests that the vehicle is more efficiently packed as number of hops increases. Since only one passenger is assigned to a vehicle in MOVI, its effective distance is 1. It is evident that MHRS is efficient as compared to DeepPool since the former achieves better overall utilization of vehicles even when the number of hops is close to 1.

Figures \ref{idle_vs_accept}, \ref{wait_vs_accept} and \ref{veh_vs_accept} plot the idle time, wait time and number of vehicles used respectively for different accept rates. We change the values of $\beta_{1}$ to obtain the figures. The values of $\beta_{1}$ used are - $5$, $7$, $8$, $9$ and $10$.

The idle time per vehicle is captured in Figure \ref{idle_vs_accept}. Idle time is defined as total time when the cars are not serving any request divided by the total number of requests. This metric gives an indication on how much time a vehicle is burning extra fuel without serving any customer or gaining revenue. Clearly, MHRS leads to a significant decrease in the idle cruising time of vehicles. Small idle time would encourage incessant revenue generation for the taxi provider firms as well as quick fulfillment of customer demands (This can also be inferred from Figure \ref{accept_vs_hop}).

Figure \ref{wait_vs_accept} shows that the wait time decreases as the accept rate increases. Wait time is defined as the average difference between the time of request for a passenger and the time when the passenger starts his/her ride.  MHRS outperforms DeepPool and MOVI, and thus suggesting more than one hop allows customers to have sufficient vehicles in their vicinity and hence less waiting time on average. 
The number of used vehicles for different accept rate is shown in Figure \ref{veh_vs_accept}.  The number of used vehicles is defined as the average number of vehicles used to fulfill the requests. We note that MHRS needs  
fewer number fo vehicles as compared to DeepPool and MOVI. This indicates that MHRS contributes towards reduced fuel consumption and less traffic. Further, it leads not only to a constant circulation of supply but also  helps in minimizing the fleet size and maintenance cost.

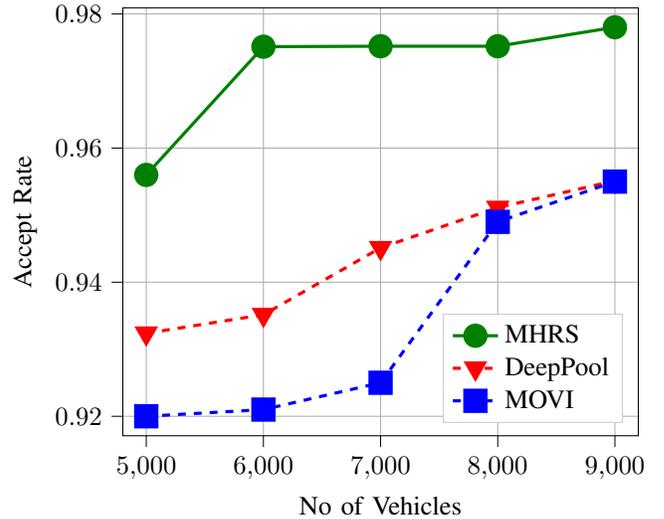
\begin{figure}[t]
% This file was created by tikzplotlib v0.8.4.
\begin{tikzpicture}

\begin{axis}[
legend cell align={left},
legend style={at={(0.97,0.03)}, anchor=south east, draw=white!80.0!black},
tick align=outside,
tick pos=left,
x grid style={white!69.01960784313725!black},
xlabel={No of Vehicles},
xmajorgrids,
xmin=4800, xmax=9200,
xtick style={color=black},
y grid style={white!69.01960784313725!black},
ylabel={Accept Rate},
ymajorgrids,
ymin=0.917099009013976, ymax=0.980920810706508,
ytick style={color=black}
]
\addplot [very thick, green!50.19607843137255!black, mark=*, mark size=4, mark options={solid}]
table {%
5000 0.955999092093317
6000 0.975119802691333
7000 0.975189802681345
8000 0.975190038386263
9000 0.978019819720484
};
\addlegendentry{MHRS}
\addplot [very thick, red, dashed, mark=triangle*, mark size=4, mark options={solid,rotate=180}]
table {%
5000 0.93237865
6000 0.93514532
7000 0.94512361
8000 0.951129087
9000 0.95511654321
};
\addlegendentry{DeepPool}
\addplot [very thick, blue, dashed, mark=square*, mark size=4, mark options={solid}]
table {%
5000 0.92
6000 0.921
7000 0.925
8000 0.949
9000 0.955
};
\addlegendentry{MOVI}
\end{axis}

\end{tikzpicture}
% \captionbox{Average accept rate for different number of vehicles.}
\caption{Average accept rate for different number of vehicles.}
		\label{accept_vs_vehicles}
\end{figure}

% \begin{figure}[t]
% 	\centering
% %	\begin{minipage}[b]
% %		{0.45\linewidth}
% 		\includegraphics[scale=1,width=0.4\textwidth]{{"new_figs_v3/accept_rate_vehicle_num_v3"}.eps}
% %		\vspace{-.3in}
% 		\caption{Average wait time for different number of vehicles.}
% 		\label{wait_vs_vehicles}
% %	\end{minipage}
% \end{figure}

\begin{figure}[t]	
%	\begin{minipage}
	\centering
%		[b]{0.52\linewidth}
		\includegraphics[scale=1,width=0.45\textwidth]{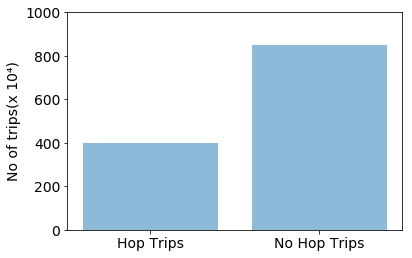}
%		\vspace{-.3in}
		\caption{Distribution of customer trips with and without hops for our proposed MHRS algorithm.}
		\label{veh_vs_hops}
%	\end{minipage}
%	\hspace{.01in}
%	\begin{minipage}[b]{0.32\linewidth}
%		\includegraphics[scale=1,width=\textwidth]{{"new_figs/heat map pickup drop"}.png}
%		\vspace{-.3in}
%		\caption{Average idle time  for different number of vehicles.}
%		\label{veh_vs_idletime}
%	\end{minipage}
\end{figure}

The effect of varying the number of  vehicles on waiting time and idle time 
is captured in Figures \ref{veh_vs_idletime} and \ref{accept_vs_vehicles}. We only change the number of vehicles and set our parameters as - $\beta_1=5$, $\beta_{2}=1$, $\beta_{3}=3.5$, $\beta_{4}=0.05$, and $\beta_{5}=2$. Intuitively, as the number of vehicles increases, the idle time increases since more vehicles will compete the existing passengers and further cruise searching for potential customers. However, our approach of MHRS still achieves the lowest idle time even in the regime of low number of vehicle. While DeepPool allows pooling customers together, it remains a nook option with limited route choice and few rides per route can be pooled. This highlights the importance of allowing customers to transfer vehicles to further improve the quality of experience of passengers. 

Figure \ref{accept_vs_vehicles} plots the accept rate for different number of vehicles.
The gain of the MHRS over both DeepPool and MOVI policies is remarkable even for low number of vehicles, which highlights the benefits of ride-sharing with passenger transfer capabilities in meeting customer demand at small travel time overhead.  We observe that both DeepPool and MOVI achieve higher accept rate as the number of vehicle increase. This is because more vehicles can be dispatched. Further, DeepPool algorithm can serve more customers using a single car by better utilizing the passengers proximity. However, this comes at an additional increased in the idle time of vehicles as depicted in Figure \ref{veh_vs_idletime}. By allowing passengers transfer, vehicles can pool even more customers which results in increasing the accept rate of passengers.    

The distribution of customer trips that encounter hops/transfers during their paths to destination is shown in Figure \ref{veh_vs_hops}. This figure also shows the total number of customer trips without hops (i.e., total number of passenger trips that reach their destinations using a single vehicle). We can see that approximately $50\%$ of the total trips involve customer transfers. By seeing this figure along with Figures \ref{accept_vs_hop}--\ref{idle_vs_accept}, we conclude that most of the passengers experience only one transfer/hop at most. Hence, with only {\it one} transfer (going from one direct trip to two hops) on $50\%$ of the total trips, we notice significant reductions in customers waiting times, vehicle idle times, and the total number of used vehicle. This shows that it would be sufficient to search only for one transfer since shared rides with two (or more) hops does not give significantly better performance, on an average. 

\section{Conclusion and Future Work}

We have looked at the problem of multi-hop ride-sharing, where
passengers can be dropped one or more times before reaching their destinations.
%passengers can ride with one vehicle for sometime and then transfer to another one.
We have proposed an efficient distributed model-free algorithm for dispatching and matching policies where
reinforcement learning techniques are used to
to learn the optimal decisions for each vehicle individually.
We have observed better accept rates, more efficient utilization of vehicles, a constant revenue cycle and reduced prices for customers. 
%As more and more autonomous vehicles, especially trucks, come in, multi hop sharing algorithm can be used to move freight and packages from one place to another more efficiently and at lesser budgets.
Further, we have seen a significant improvement in waiting time, idle time, and the number of used vehicles when going from one hop to two hops, but we
have not seen any considerable changes when further increasing
the hop count. This indicates that searching for shared rides with three (or more) hops does not give significantly better results, on an average. Surprisingly, a negligible improvement is noticed for more than two hops.

As a future research, utilizing the interaction between vehicles in a multi-agent reinforcement learning settings is a nice future direction.
Further, optimal incentive techniques to
influence passengers to opt for the MHRS is another promising research area. Lastly, as a promising alternative to the inefficient traditional package delivery techniques, MHRS systems represent a potential delivery capability through ridesharing within an urban environment. By allowing the
passengers and the packages with similar itineraries and time schedules to share one vehicle, the delivery cost and time can be reduced significantly.

%As more and more autonomous vehicles, especially trucks, come in, multi hop sharing algorithm can be used to move freight and packages from one place to another more efficiently and at lesser budgets.

% \end{document}

%\clearpage
%\newpage
%\small
\bibliographystyle{IEEEtran}
%beurocratic\bibliographystyle{acm}

\bibliography{bib}

\end{document}